\documentclass{article} % For LaTeX2e

\usepackage{nips14submit_e,times}
\usepackage{moreverb,url}
\usepackage[pdftex]{graphicx}
\usepackage{amsmath}
\usepackage{amssymb}
\usepackage[justification=centering]{caption}
\usepackage{graphicx}
\usepackage{amssymb}
\usepackage{amsthm}
\usepackage{amsmath}
\usepackage{mathtools}
\usepackage{caption}
\usepackage{breqn}
\usepackage{cite, bm}
\usepackage{float}
\usepackage{algpseudocode}
\usepackage{algorithm}

\usepackage{xfrac}
\usepackage{enumitem}
\usepackage{subcaption}
\usepackage{makebox}
\usepackage{multirow}
\usepackage{longtable}
\usepackage[outdir=./]{epstopdf}
\usepackage[export]{adjustbox}
\usepackage{soul,color}
\usepackage{fancyhdr}
\usepackage{hyperref}
\usepackage{makebox}
\usepackage{array}

\graphicspath{{images/}}

\nipsfinalcopy % Uncomment for camera-ready version

\begin{document}

\title{On Correlation of Features Extracted by Deep Neural Networks}

\author{Babajide~O. Ayinde\thanks{B.~O.~Ayinde is with the Department
		of Electrical and Computer Engineering, University of Louisville, Louisville,
		KY, 40292 USA (e-mail: babajide.ayinde@louisville.edu).}
     , Tamer Inanc\thanks{T.~Inanc is with the Department
		of Electrical and Computer Engineering, University of Louisville, Louisville,
		KY, 40292 USA, (e-mail: t.inanc@louisville.edu).}% <-this % stops a space
	, and Jacek~M.~Zurada\thanks{J.~M.~Zurada is with the Department
		of Electrical and Computer Engineering, University of Louisville, Louisville,
		KY, 40292 USA, and also with the Information Technology Institute, University of Social Science,\L \'{o}dz 90-113, Poland (Corresponding author, e-mail: jacek.zurada@louisville.edu). This work was supported by the NSF under grant 1641042.}% <-this % stops a space
}

\maketitle

\begin{abstract}
Redundancy in deep neural network (DNN) models has always been one of their most intriguing and important properties. DNNs have been shown to overparameterize, or extract a lot of redundant features. In this work, we explore the impact of size (both width and depth), activation function, and weight initialization on the susceptibility of deep neural network models to extract redundant features. To estimate the number of redundant features in each layer, all the features of a given layer are hierarchically clustered according to their relative cosine distances in feature space and a set threshold. It is shown that both network size and activation function are the two most important components that foster the tendency of DNNs to extract redundant features. The concept is illustrated using deep multilayer perceptron and convolutional neural networks on MNIST digits recognition and CIFAR-10 dataset, respectively.
\end{abstract}


\section{Introduction}
DNNs have become ubiquitous in a wide range of applications ranging from computer vision \cite{simonyan2014very,krizhevsky2012imagenet,he2016deep,ayinde2018deep} to speech recognition \cite{graves2013speech,graves2014towards,oord2016wavenet} and natural language processing \cite{jozefowicz2016exploring,shazeer2017outrageously}. Over the past few years, the general trend has been that DNNs have grown deeper and wider, amounting to huge increase in their size. The number of parameters in DNNs is usually very large and no constraints are generally placed on the data and/or the model, hence offering possibility to learn very flexible and high-performing models \cite{liu2014pruning,ayinde2018building}. However, this flexibility may hinder their scalability and practicality due to very high memory/time requirements, and may lead to extracting highly redundant parameters with risk of over-fitting \cite{yoon2017combined,ayinde2019regularizing}.\\
\indent
A number of studies have shown that a significant percentage of features extracted by DNNs are redundant \cite{denil2013predicting,rodriguez2016regularizing,bengio2009slow,changpinyo2017power,ayinde2017nonredundant,han2015deep,han2016dsd,ayinde2016clustering}. As demonstrated in \cite{denil2013predicting}, a fraction of the parameters is sufficient to reconstruct the entire network by simply training on low-rank decompositions of the weight matrices. To this end, Optimal Brain Damage \cite{NIPS1989_250} and Optimal Brain Surgeon \cite{hassibi1993second} exploit second-order derivative information of the loss function to localize unimportant parameters. HashedNets use a hash function to randomly group weights into hash buckets, so that all weights within the same hash bucket share a single parameter value for pruning purposes \cite{chen2015compressing}. Redundant features have also been localized and pruned using simple thresholding mechanism \cite{han2015learning}. \\
\indent
Instead of localizing the redundant neurons in a fully-connected network, \cite{mariet2015diversity} compresses a trained model by identifying a subset of diverse neurons. Redundant feature maps are removed from a well trained network using particle filtering to select the best combination from a number of randomly generated masks \cite{anwar2017structured}. The importance of features has also been ranked based on the sum of their absolute weights \cite{li2016pruning}. With the assumptions that features are co-dependent within each layer, \cite{ioannou2016deep} groups features in hierarchical order. Driven by feature map redundancy, \cite{zhang2016accelerating} factorizes a layer into $3\times3$ and $1\times1$ combinations and prunes redundant feature maps.\\
\indent
These observations that DNNs are prone to extracting redundant features evidently suggest and reinforce our hypothesis that much of the information stored within DNN models may be redundant. In addition, large number of parameters also translates to model's tendency to overfitting, if trained on limited amount of data. Some of the problems associated with over-parameterization have been previously addressed via model compression \cite{li2016pruning,polyak2015channel}, removal of unnecessary weights \cite{NIPS1989_250,hassibi1993second}, and regularization \cite{zhou2016less,wen2016learning}. These heuristics for eliminating redundancy more often than not deteriorate the performance of the compressed model. The open question still remains: how to obtain best compact and efficient models that are free of redundancy? \\
\indent
Knowing the level of redundancy in models could be useful for two main reasons. First, inference-cost-efficient models can be built via pruning with small deterioration of prediction accuracy \cite{li2016pruning,han2015learning}. This is important in practice because optimal architecture are unknown. However, pruning should enable smaller model to inherit knowledge from a larger model. Since learning a complex function directly by small suboptimal model might result in its poor performance, it is therefore necessary to first learn a task with model many parameters and to follow with pruning redundant and less important features \cite{anwar2017structured}. This is particularly important for porting deep learning models to resource limited portable devices. Secondly, information about the level of redundancy in models can be used for feature diversification in order to optimize their performance since the adverse effect of redundancy in DNNs has been shown in \cite{xie2015generalization,rodriguez2016regularizing,rodriguez2016regularizing,yoon2017combined,cogswell2015reducing}.\\
\indent
The problem addressed in this work is three-fold: (i) we investigate the impact of modules of DNNs such as width and depth of the network, activation function, and parameter initialization on extraction of redundant features by DNNs (both fully-connected and convolutional), (ii) we estimated the number of redundant features by adapting hierarchical agglomerative clustering algorithm, and (iii) we show experimentation in order to obtain insight about which configurations (network size/activation function/parameter initialization) provide better performance tradeoff. The paper is structured as follows: Section
II introduces the network configurations and the notation used in the paper. Section III introduces the notion of feature redundancy in DNN and its estimation. Section IV discusses the experimental designs and present the results. Finally, conclusions are drawn in the last section.
\section{Network Configurations}
Notations and network configurations in the paper in context of convolutional and fully-connected layer are briefly and separately highlighted below:\\
\subsubsection{Convolutional layer:}
By letting $n'_l$, $h_l$, and $w_l$, denote the number of channels, height and width of input of the $l^{th}$ layer, respectively, input $x_l \in \mathbb{R}^{p}$ is transformed by a layer into output $x_{l+1} \in \mathbb{R}^{q}$, where $x_{l+1}$ serves as the input in layer $l+1$. Since the layer is convolutional, $p$ and $q$ are given as $n'_l \times h_l \times w_l$  and $n'_{l+1}\times h_{l+1}\times w_{l+1}$, respectively. A convolutional layer convolves $x_l$ with $n'_{l+1}$ $3D$ features $\chi \in \mathbb{R}^{n'_l \times k\times k}$, resulting in $n'_{l+1}$ output feature maps. Each $3D$ feature consists of $n'_l$ $2D$ kernels $\zeta \in k\times k$. Unrolling and combining all features into a single kernel matrix $\mathbf{W} \in \mathbb{R}^{z\times n'_{l+1}}$ where $z= k^2n'_l$. The $ith$ feature in layer $l$ is denoted by $\mathbf{w}^{(l)}_i$, i=1,...$n'_l$ and each $\mathbf{w}^{(l)}_i \in \mathbb{R}^{z}$ corresponds to the $i$-th column of the kernel matrix $\mathbf{W}^{(l)} = [\mathbf{w}^{(l)}_1, \;\;...\mathbf{w}^{(l)}_{n'_l}] \in \mathbb{R}^{z\times n'_{l+1}}$.\\
\subsubsection{Fully-connected layer:}
In the case of a fully-connected layer, $p$ and $q$ denote $n'_lh_lw_l \times 1$ and $n'_{l+1} \times 1$, respectively. A layer operation involves only vector-matrix multiplication with kernel matrix $\mathbf{W} \in \mathbb{R}^{z\times n'_{l+1}}$, where $z=n'_lh_l w_l$. Also for fully-connected layer, the $ith$ feature in layer $l$ is denoted by $\mathbf{w}^{(l)}_i$, i=1,...$n'_l$ and each $\mathbf{w}^{(l)}_i \in \mathbb{R}^{z}$ corresponds to the $i$-th column of the kernel matrix $\mathbf{W}^{(l)} = [\mathbf{w}^{(l)}_1, \;\;...\mathbf{w}^{(l)}_{n'_l}] \in \mathbb{R}^{z\times n'_{l+1}}$.
\section{Estimating the number of redundant features}
Correlation between two features can be computed by evaluating the cosine similarity measure between them as given in \eqref{MyEq1}:
\begin{equation} \label{MyEq1}
  \begin{split}
  Cosine(\phi_1,\phi_2) = \frac{<\phi_1,\phi_2>}{\parallel\phi_1\parallel\parallel\phi_2\parallel}
  \end{split}
\end{equation}
where $<\phi_1,\phi_2>$ is the inner product of two arbitrary normalized feature vectors $\phi_1$ and $\phi_2$; $\phi_i= \sfrac{w_i}{\sqrt{||w_i||^2}}$ and $i=1,2$. The similarity between two feature vectors corresponds to the correlation between them, that is, the cosine of the angle between them in feature space. Since the entries of feature vectors can take both negative and positive values, $Cosine(\phi_1,\phi_2)$ is bounded by [-1,1]. It is 1 when $\phi_1$=$\phi_2$ or when $\phi_1$ and $\phi_2$ are identical. $Cosine(\phi_1,\phi_2)$ is -$1$ when the two vectors are in exact opposite direction. The two feature vectors are orthogonal in weight space when $Cosine$ is $0$.\\
\indent
The evaluation of pairwise feature similarities $\Omega^{(l)}$ for a given layer $l$ can be vectorized to reduce the computational overhead. By letting $\mathbf{\Phi}=[\phi^{(l)}_1, \;\;...\phi^{(l)}_{n'_l}] \in \mathbb{R}^{z\times n'_l}$ contain $n'_l$ normalized feature vectors $\phi_i$ as columns, each with $z$ elements corresponding to connections from layer $l-1$ to $i^{th}$ neuron of layer $l$, then the pairwise feature similarities $\Omega^{(l)}$ for a given layer $l$ is given as
\begin{equation} \label{MyEq1b}
  \begin{split}
  \Omega^{(l)} = \mathbf{\Phi^T}^{(l)}\mathbf{\Phi}^{(l)}
  \end{split}
\end{equation}
$\Omega^{(l)} \in \mathbb{R}^{n'\times n'}$ contains the inner products of each pair of columns $i$ and $j$ of $\mathbf{\Phi}^{(l)}$ in each position $i$,$j$ of $\Omega$ in layer $l$. It is remarked that $\Omega^{(l)}$ can be used to roughly estimate the number of redundant features in layer $l$. In this work, we utilize a suitable agglomerative similarity testing/clustering algorithms to estimate the number of redundant features. \\
\indent
Based on a comparative review, a clustering approach from \cite{walter2008fast,ding2002cluster} has been adapted and reformulated for this purpose. By starting with each feature vector $\phi_i$ as a potential cluster, agglomerative clustering is performed by merging the two most similar clusters $C_a$ and $C_b$ as long as the average similarity between their constituent feature vectors is above a chosen cluster similarity threshold denoted as $\tau$ \cite{leibe2004combined,manickam2000intelligent}. The pair of clusters $C_a$ and $C_b$ exhibits average mutual similarities as follows:
\begin{equation} \label{MyEq11}
  \begin{split}
  \overline{SIM_C}(C_a, C_b) &= \frac{\sum_{\phi_i\in C_a, \phi_j\in C_b}Cosine(\phi_i,\phi_j)}{|C_a|\times |C_b|} > \tau \\
   & a,b = 1,...n'_l;\,\, a\neq b ;\,\,\,  i = 1,...|C_a|; \\
   & j=1,...|C_b|;  \;\;\;\  and \;\;\;\; i\neq j
  \end{split}
\end{equation}
\\
\indent
It must be noted that the definition of similarity in \eqref{MyEq11} uses the graph-based-group-average technique, which defines cluster proximity/similarity as the average of pairwise similarities of all pairs of features from different clusters. This work also considers other similarity definitions such as the single and complete links. Single link defines cluster similarity as the similarity between the two closest feature vectors that are in different clusters. On the other hand, complete link assumes that cluster proximity is the proximity between the two farthest feature vectors of different clusters. In this work, experiments based on average proximity were reported because of their superior performance. It is strongly believe that graph-based-group-average performs better because all cluster members contributed in the decision making process.\\
\indent
The objective of the clustering algorithm is to discover $n_f$ features in the set of $n'$ original weight vectors (or simply features), where $n_f \leq n'$. Upon detecting these  distinct $n_f$ clusters, a representative feature from each of these $n_f$ clusters is randomly sampled without replacement and the remaining features in that cluster are tagged as redundant. The number of redundant features in a particular layer $l$ is then estimated as in \eqref{MyEq12}.
\begin{equation} \label{MyEq12}
  \begin{split}
  n_r^{(l)} &= n'_l - n_f^{(l)}
  \end{split}
\end{equation}
\begin{figure*}[htb!]
	\begin{minipage}[b]{0.5\linewidth}
		\centering
		\centerline{\includegraphics[scale=0.4]{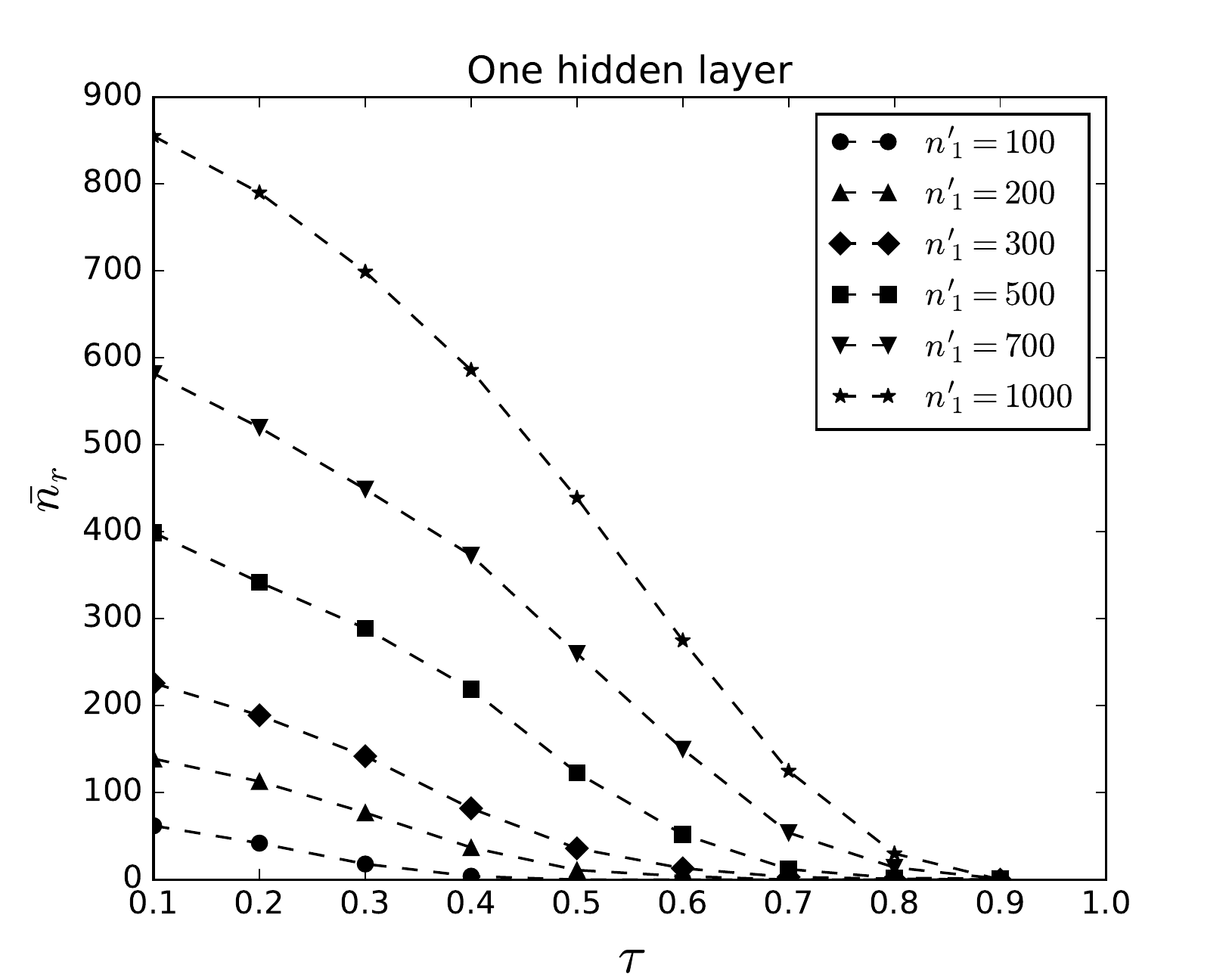}}%norb_subset
		{{\footnotesize (a)}}
	\end{minipage}
\begin{minipage}[b]{0.5\linewidth}
		\centering
		\centerline{\includegraphics[scale=0.4]{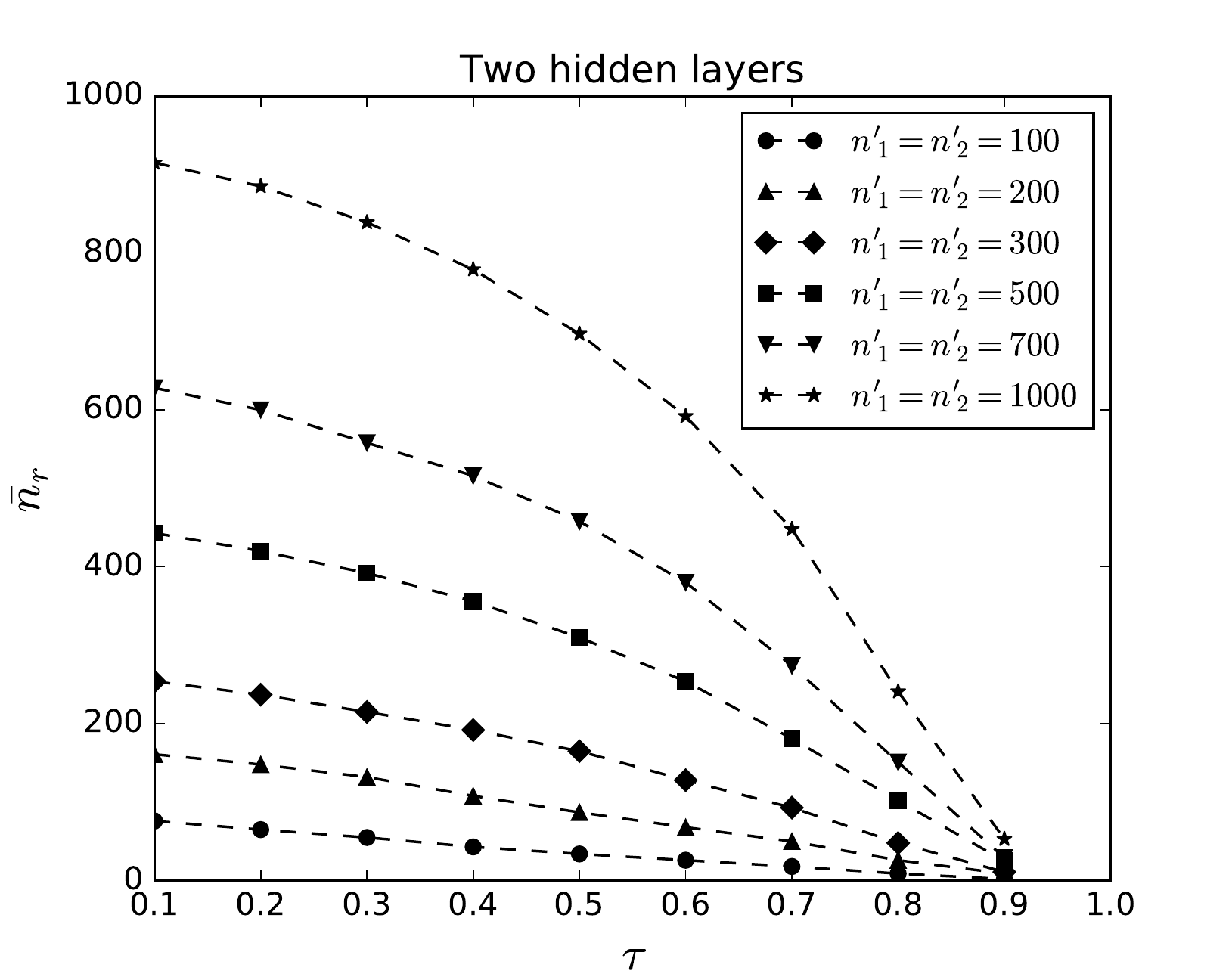}} %
		{{\footnotesize(b)}}
	\end{minipage}
\begin{minipage}[b]{0.5\linewidth}
		\centering
		\centerline{\includegraphics[scale=0.4]{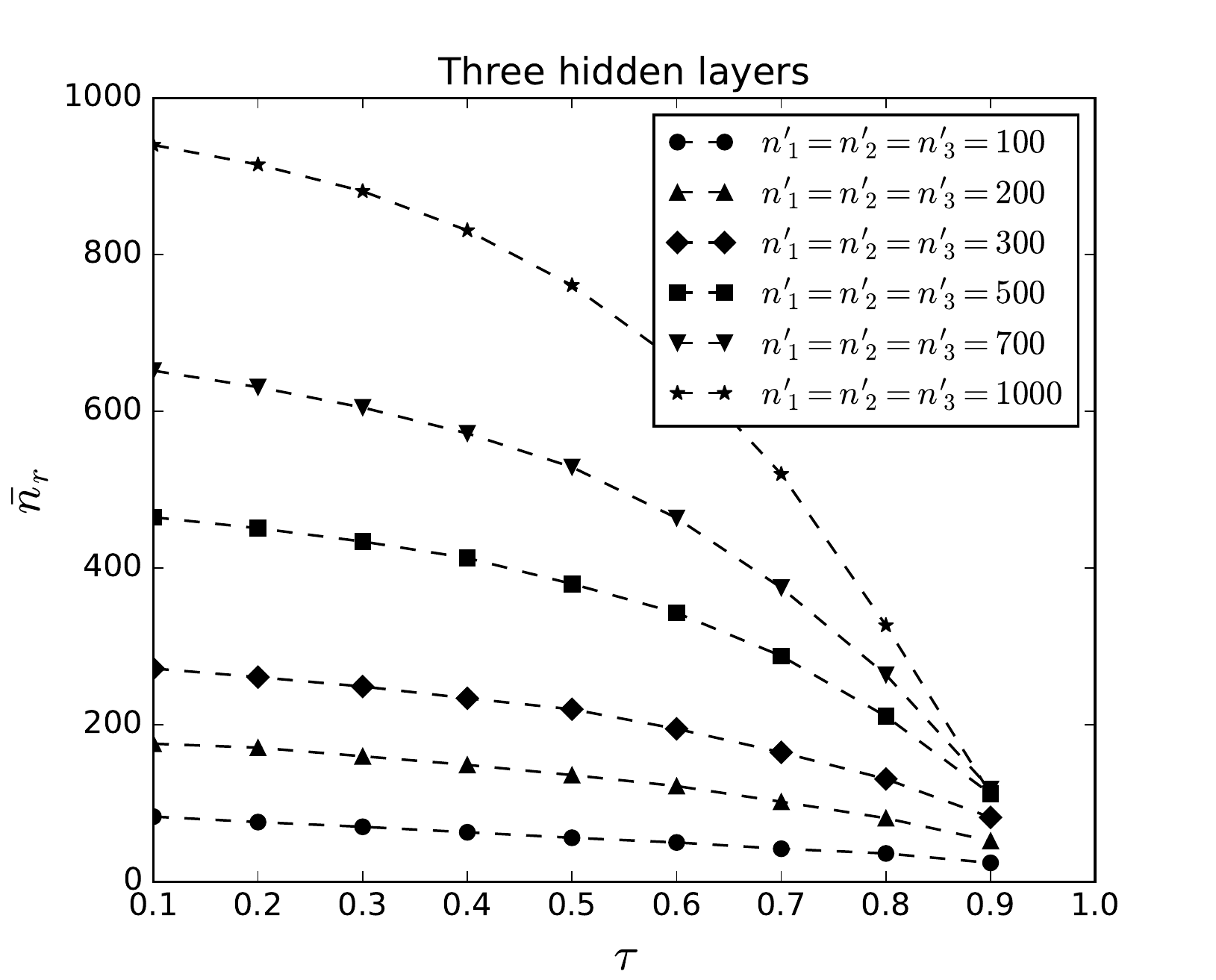}} %
		{{\footnotesize(c)}}
	\end{minipage}
\begin{minipage}[b]{0.5\linewidth}
		\centering
		\centerline{\includegraphics[scale=0.4]{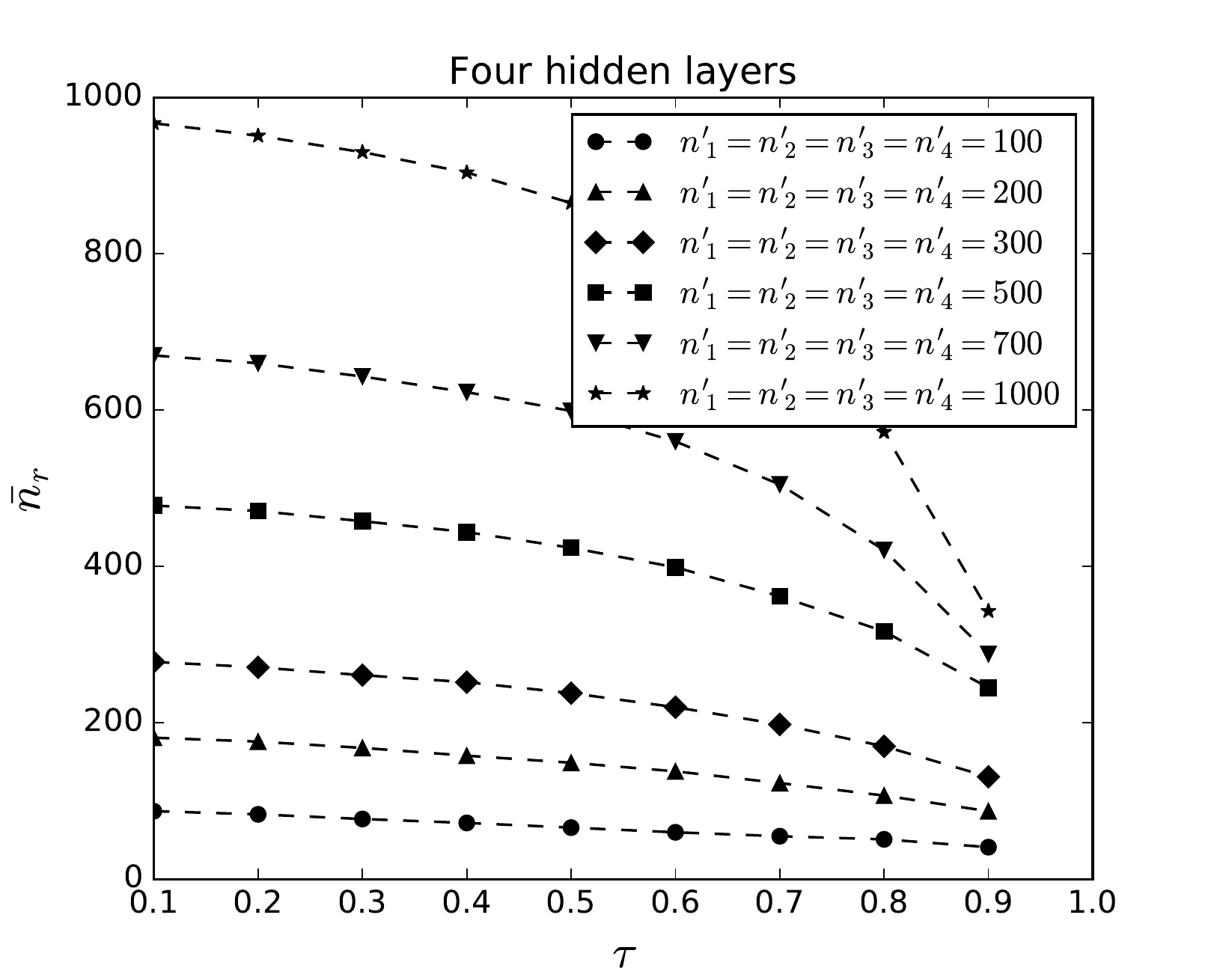}} %
		{{\footnotesize(d)}}
	\end{minipage}
	\caption{Average number of redundant features across all layers ($\bar{n}_r$) vs threshold $\tau$ with (a) one (b) two (c) three, and (d) four hidden layers of multilayer percerptron using MNIST dataset. Network width is the number of hidden units per layer and network depth is the number of hidden layers. Networks with more than one hidden layer have equal number of hidden units in all layers.}\label{fig1}
\end{figure*}
\begin{figure*}[htb!]
\centering
%\captionsetup{justification=centering,margin=2cm}
	\begin{minipage}[b]{0.32\linewidth}
		\centering
       %\captionsetup{justification=centering}
		\centerline{\includegraphics[scale=0.35]{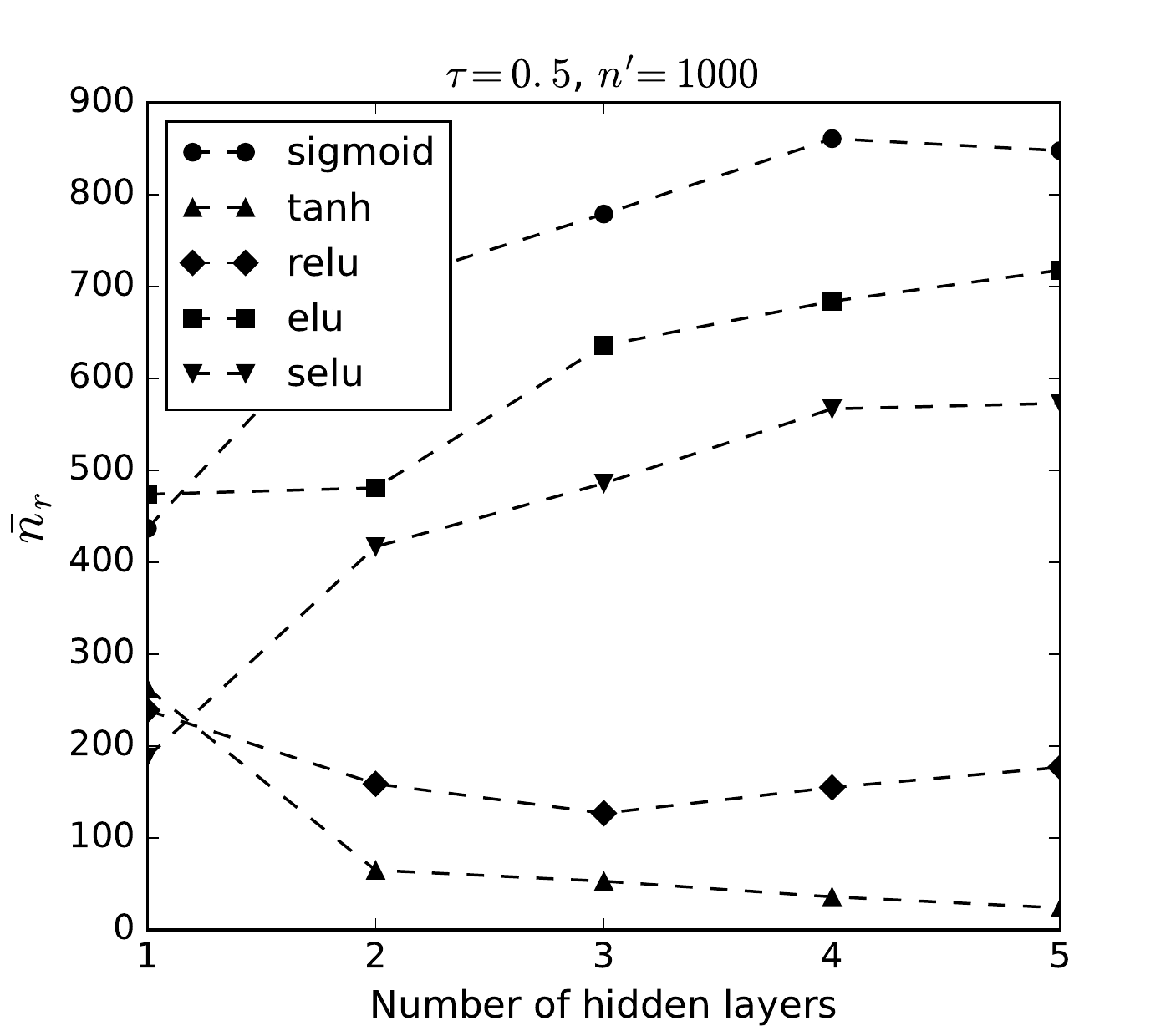}}
		{{\footnotesize (a)}}
	\end{minipage}
	%%\hfill
	\begin{minipage}[b]{0.32\linewidth}
		\centering
		\centerline{\includegraphics[scale=0.35]{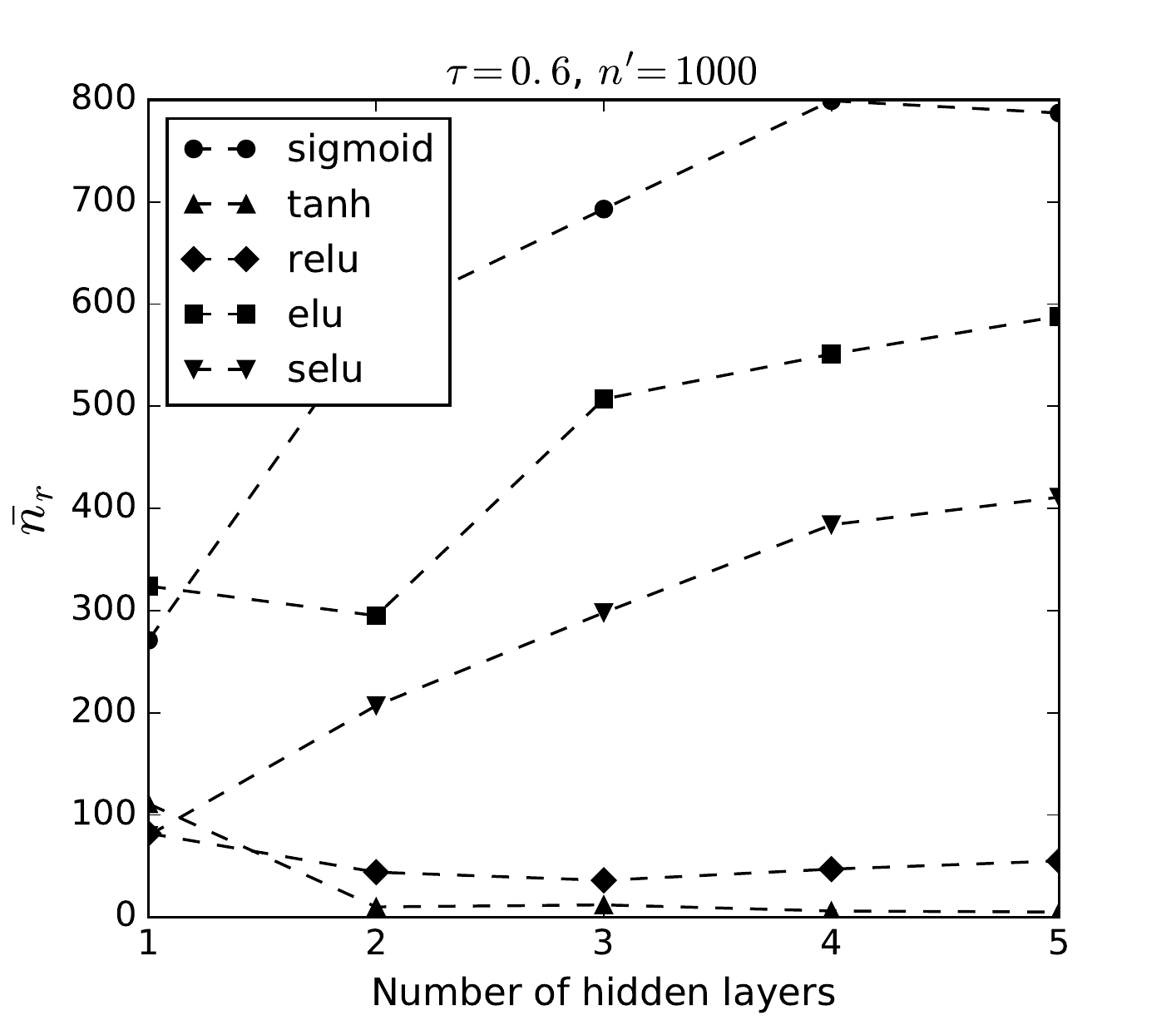}}%, height=5cm
		{{\footnotesize(b)}}
    \end{minipage}
    %%\hfill
	\begin{minipage}[b]{0.32\linewidth}
		\centering
		\centerline{\includegraphics[scale=0.35]{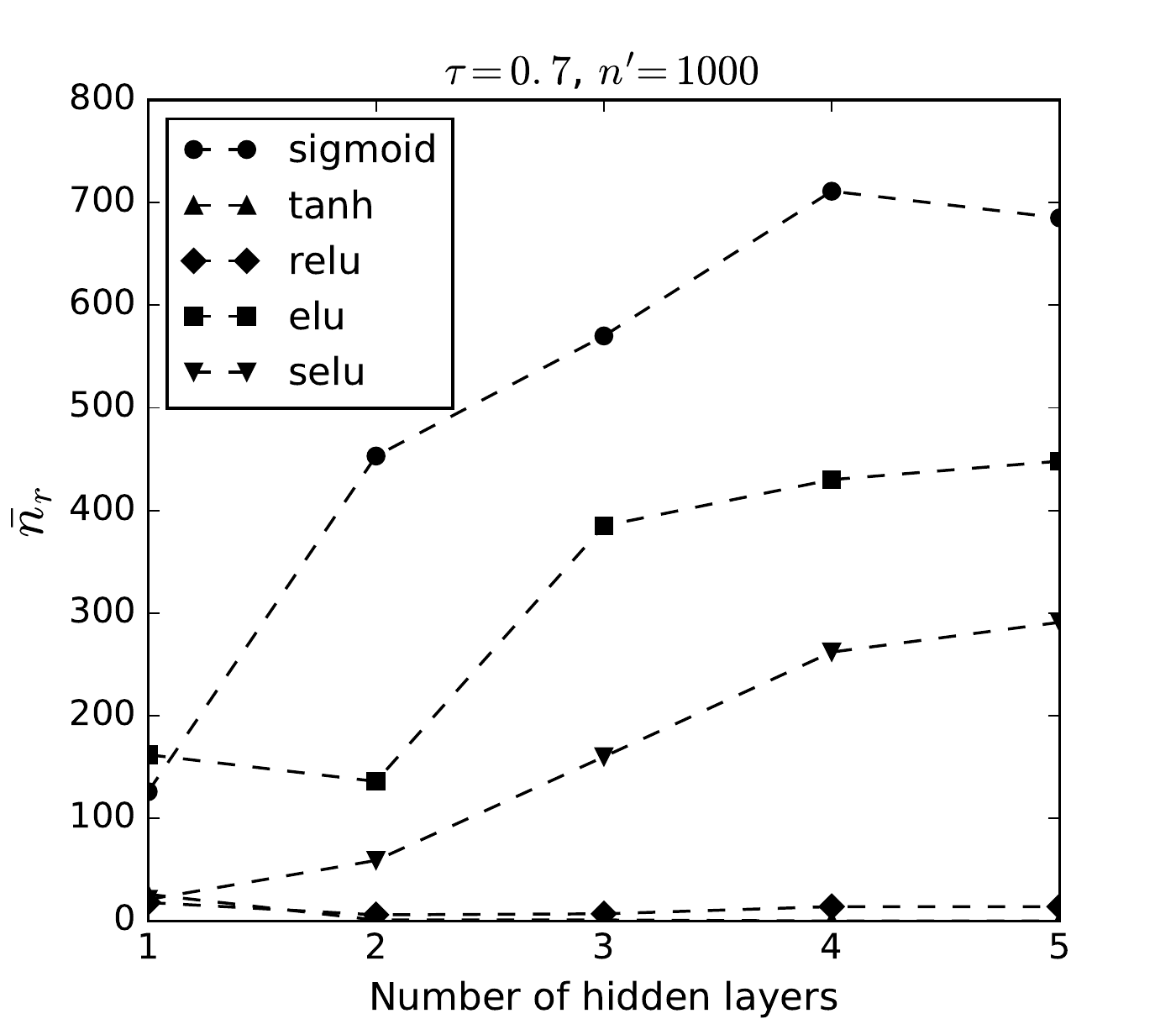}}%, height=5cm
		{{\footnotesize(c)}}
    \end{minipage}
	\caption{Average number of redundant features vs number of hidden layers of multilayer percerptron for four activation functions with (a) $\tau=0.5$ (b) $\tau=0.6$ (c) $\tau=0.7$ using MNIST dataset. Number of hidden units ($n'$) per layer is $1000$}\label{fig2}		
\end{figure*}
To illustrate the impact of activation functions and weight initializations on DNNs' susceptibility to extracting redundant features, we considered five popular activation functions (namely: Sigmoid, Tanh, ReLU, ELU, and SeLU) and five weight initializations. The activation functions considered are briefly highlighted below. Given any finite dimensional vector $\mathbf{z}$ for which activation function is to be computed for, the following four activation functions are defined as follows:
\begin{enumerate}
\item Sigmoid:
\begin{equation}\label{MyEq13}
\sigma(\mathbf{z}) = \frac{1}{1+e^{-\mathbf{z}}}
\end{equation}

\item Hyperbolic Tangent:
\begin{equation}\label{MyEq13}
Tanh(\mathbf{z}) = \frac{e^{\mathbf{z}}-e^{-\mathbf{z}}}{e^{\mathbf{z}}+e^{-\mathbf{z}}}
\end{equation}

\item Rectified Linear Units \cite{nair2010rectified}:
\begin{equation}\label{MyEq14}
\begin{split}
ReLU(\mathbf{z}) = \text{max}(0,\mathbf{z})
\end{split}
\end{equation}

\item Exponential Linear Units \cite{clevert2016fast}:
\begin{equation}\label{MyEq15}
%\begin{split}
\small{
ELU(\mathbf{z}) = \Bigg\{
\begin{array}{l l}
\mathbf{z} & \quad \mathbf{z} > 0 \\ \\
 \alpha(e^{\mathbf{z}}-1) & \quad ||\mathbf{z}|| \leq 0
\end{array}}
%\end{split}
\end{equation}
where $\alpha$ is an hyperparameter that controls the value at which ELU activation function saturates for inputs with negative values.

\item Scaled Exponential Linear Units \cite{klambauer2017self}:
\begin{equation}\label{MyEq16}
%\begin{split}
\small{
SeLU(\mathbf{z}) = \lambda \Bigg\{
\begin{array}{l l}
\mathbf{z} & \quad \mathbf{z} > 0 \\ \\
 \alpha e^{\mathbf{z}} - \alpha & \quad ||\mathbf{z}|| \leq 0
\end{array}}
%\end{split}
\end{equation}
where $\lambda>1$ and $\alpha$ are derived from the input. SeLU also uses a custom weight initialization with zero mean and standard deviation of $\sqrt{\sfrac{1}{\text{size of input vector}}}$.
\end{enumerate}
\indent
Also, the five popular weight initialization heuristics considered are briefly described as follows:
\begin{enumerate}
\item random\_uniform: initializes weights between $-0.05$ and $0.05$ from a uniform distribution

\item orthogonal \cite{saxe2013exact}: initializes weight through the generation of orthogonal matrix with scalar gain factor $g$ (chosen to be 1.0 in our experiments), where gain is a multiplicative factor that scales the orthogonal matrix

\item xavier \cite{glorot2010understanding}: is also known as Xavier uniform initialization and it initializes weights within [-$\kappa$ $\kappa$] from uniform distribution where $\kappa$ is given as:
    \begin{equation}\label{MyEq17}
    \kappa = \sqrt{\frac{6}{n'_l + n'_{l+1}}}
    \end{equation}
    where $n'_l$ is the number of input units and $n'_{l+1}$ is the number of output units.

\item he\_normal \cite{he2015delving}: initializes weight with samples drawn from a truncated normal distribution centered around $0$ with standard deviation of $\sqrt{\frac{2}{n'_l}}$

\item lecun\_normal \cite{lecun1998efficient}: initializes weight with samples drawn from a truncated normal distribution centered around $0$ with standard deviation of $\sqrt{\frac{1}{n'_l}}$.
\end{enumerate}
%This layer consists in a linear transformation of a high-dimensional input signal to ahigh-dimensional output signal with a large dense matrix defining the transformation.
\section{Experiments}
In the first set of experiments we considered a fully-connected network trained and evaluated on MNIST dataset of handwritten digits. All experiments were performed on Intel(r) Core(TM) i7-6700 CPU @ 3.40Ghz and a 64GB of RAM running a 64-bit Ubuntu 14.04 edition. The software implementation has been in Pytorch library \footnote{\url{http://pytorch.org/}} on two Titan X 12GB GPUs and the feature clustering was implemented in SciPy ecosystem \cite{scipy}. The standard MNIST dataset has 60000 training and 10000 testing examples. Each example is a grayscale image of an handwritten digit scaled and centered in a 28 $\times$ 28 pixel box. Adam optimizer \cite{kingma2014adam} with batch size of 128 was used to train the model for 200 epochs. \\
\begin{figure*}[htb!]
\centering
%\captionsetup{justification=centering,margin=2cm}
	\begin{minipage}[b]{0.32\linewidth}
		\centering
%       \captionsetup{justification=centering}
		\centerline{\includegraphics[scale=0.3]{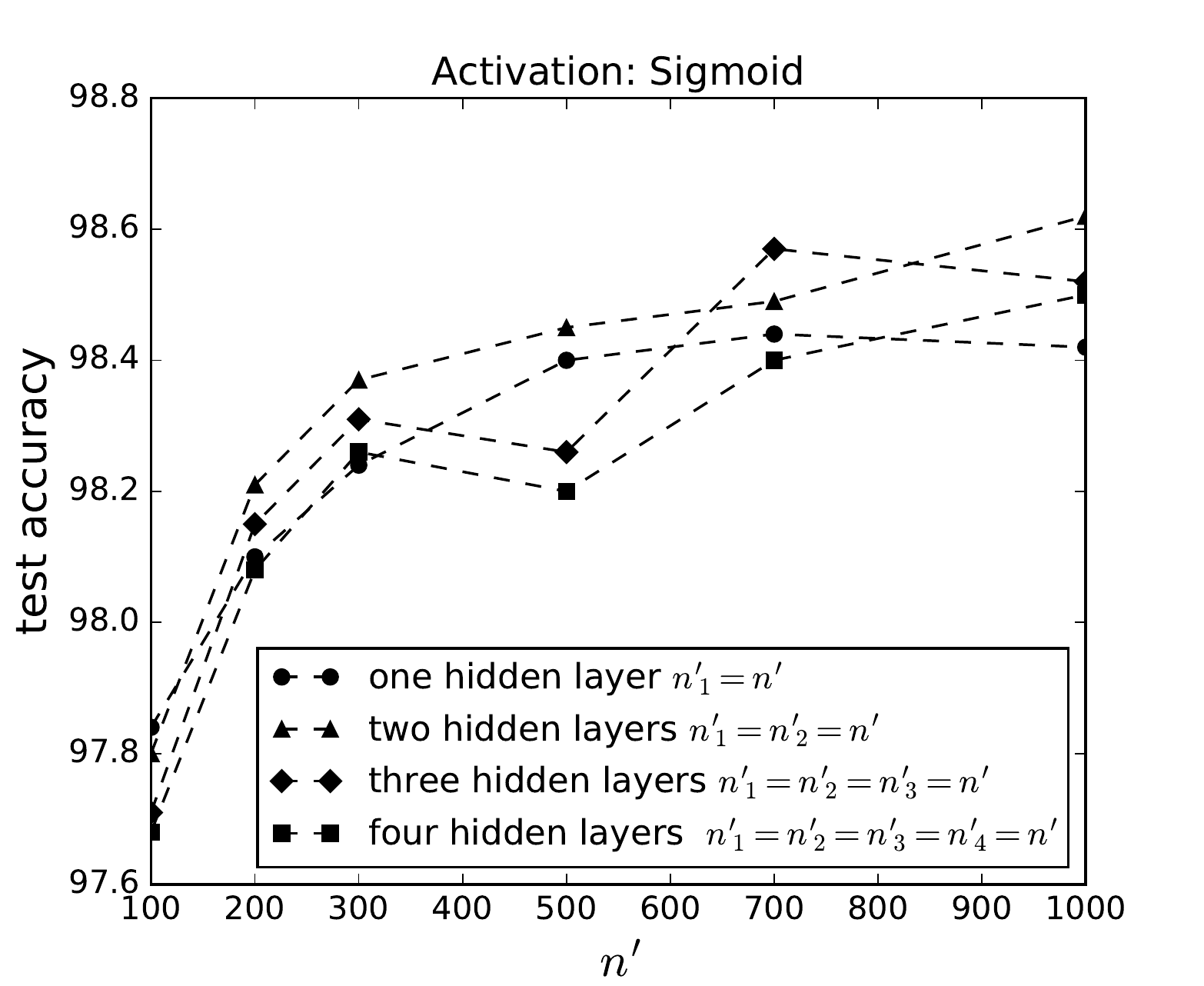}}
		{{\footnotesize (a)}}
	\end{minipage}
	%%\hfill
    \begin{minipage}[b]{0.32\linewidth}
		\centering
       \captionsetup{justification=centering}
		\centerline{\includegraphics[scale=0.3]{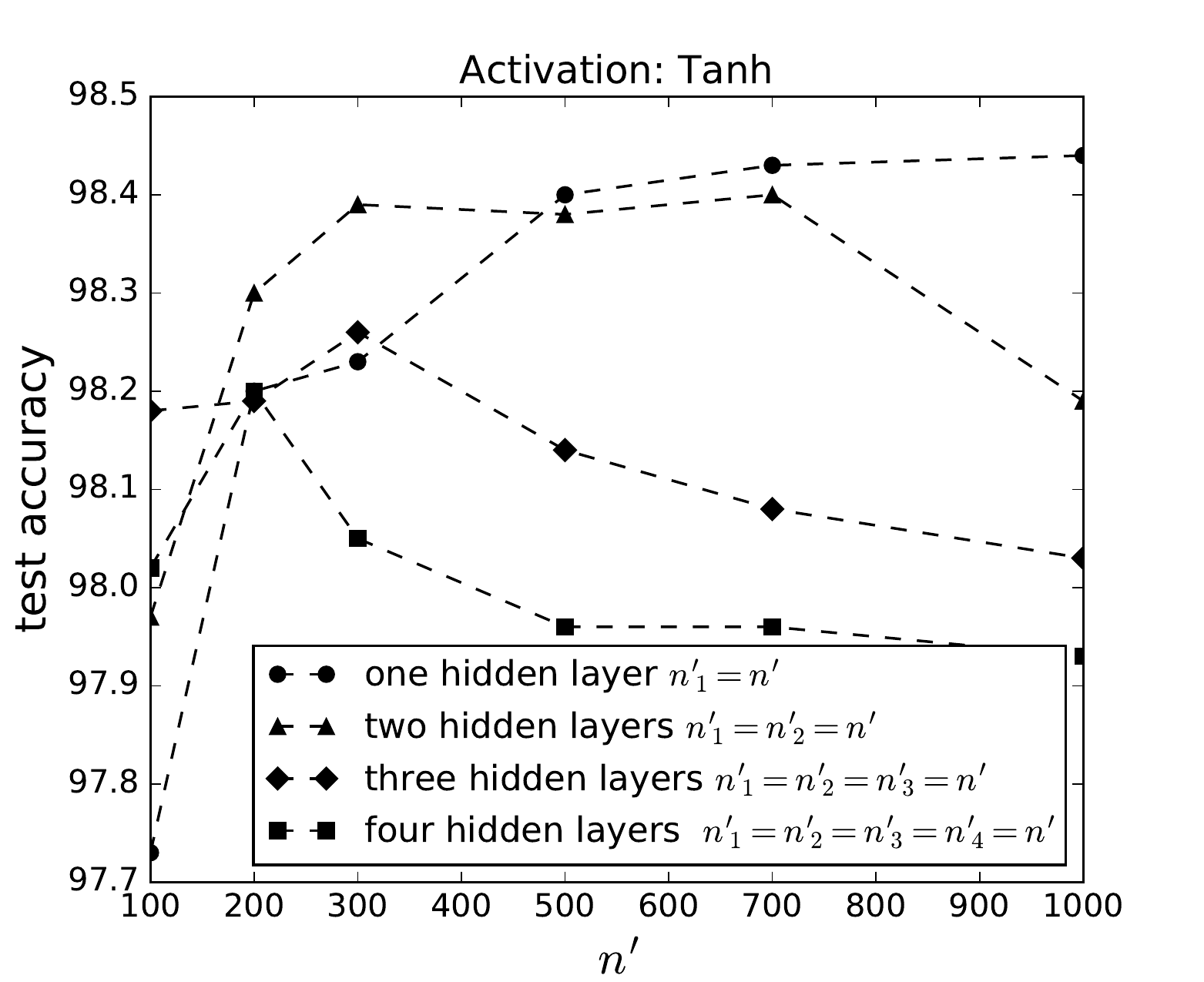}}
		{{\footnotesize (b)}}
	\end{minipage}
	%%\hfill
	\begin{minipage}[b]{0.32\linewidth}
		\centering
		\centerline{\includegraphics[scale=0.3]{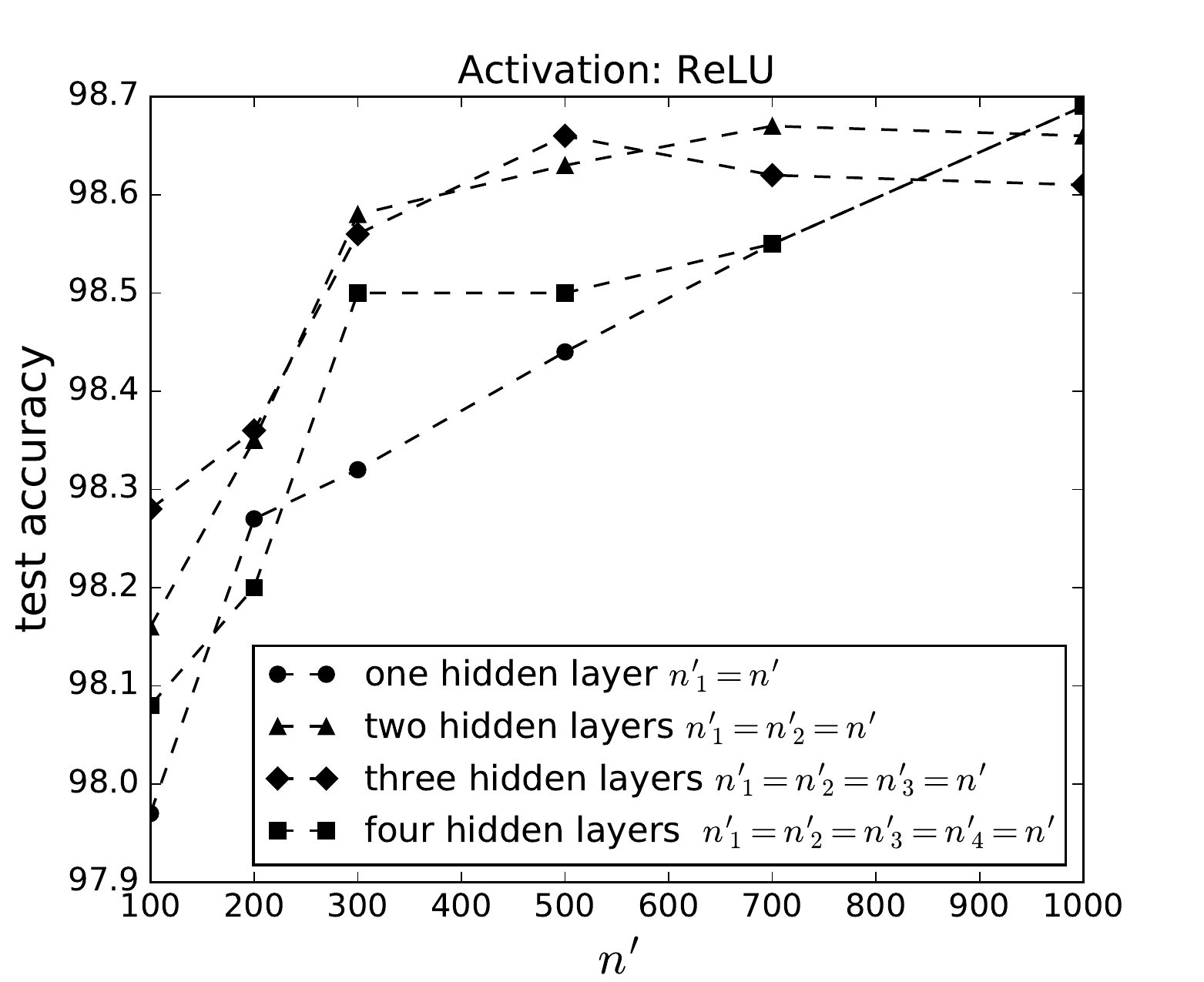}}%, height=5cm
		{{\footnotesize(c)}}
    \end{minipage}
	\caption{Performance on test set of MNIST dataset using (a) Sigmoid (b) Tanh (c) ReLU activation function}\label{fig4}		
\end{figure*}
\begin{figure*}[htb!]
\centering
%\captionsetup{justification=centering,margin=2cm}
	\begin{minipage}[b]{0.45\linewidth}
		\centering
%       \captionsetup{justification=centering}
		\centerline{\includegraphics[scale=0.4]{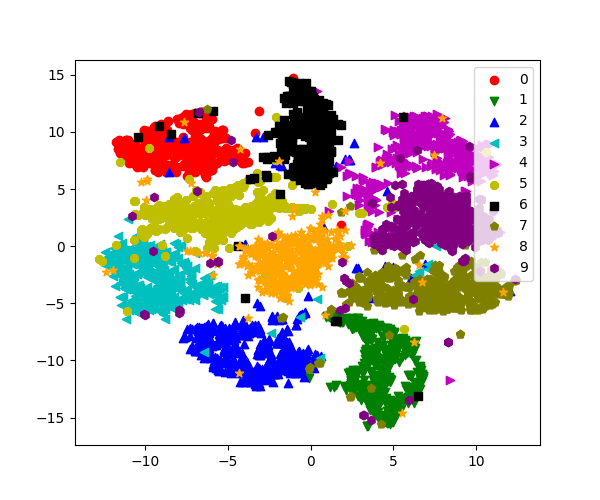}}
		{{\footnotesize (a)}}
	\end{minipage}
	%%\hfill
	\begin{minipage}[b]{0.45\linewidth}
		\centering
		\centerline{\includegraphics[scale=0.4]{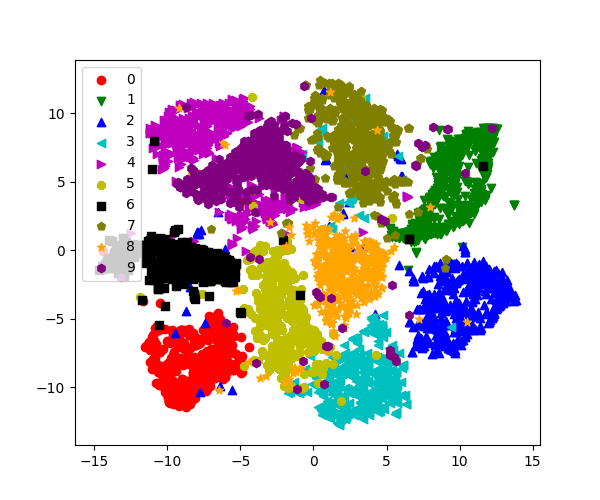}}%, height=5cm
		{{\footnotesize(c)}}
    \end{minipage}
    \begin{minipage}[b]{0.45\linewidth}
		\centering
		\centerline{\includegraphics[scale=0.4]{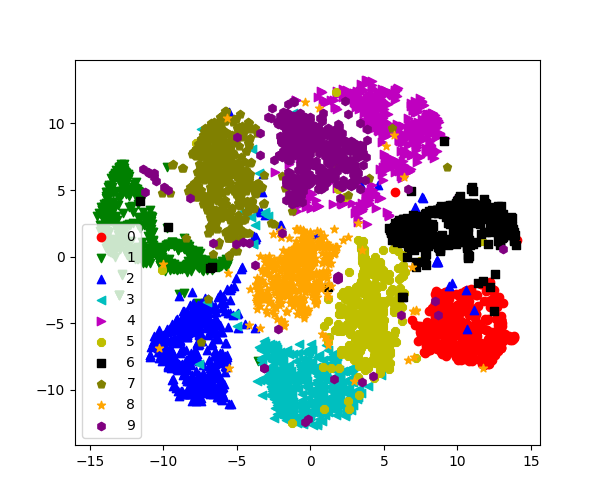}}%, height=5cm
		{{\footnotesize(c)}}
    \end{minipage}
        \begin{minipage}[b]{0.45\linewidth}
		\centering
       \captionsetup{justification=centering}
		\centerline{\includegraphics[scale=0.4]{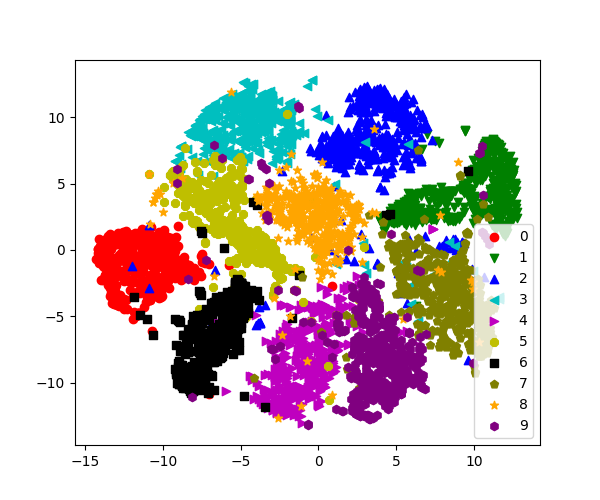}}
		{{\footnotesize (b)}}
	\end{minipage}
	\caption{t-SNE projection \cite{maaten2008visualizing} of the hidden activation of multilayer perceptron using (a) Sigmoid (b) SeLU (c) Tanh (d) ReLU activation function trained on 5000 MNIST handwritten digits test samples. All Networks have 1000 hidden units}\label{fig44}		
\end{figure*}
\begin{figure*}[htb!]
\centering
%\captionsetup{justification=centering,margin=2cm}
	\begin{minipage}[b]{0.45\linewidth}
		\centering
%       \captionsetup{justification=centering}
		\centerline{\includegraphics[scale=0.4]{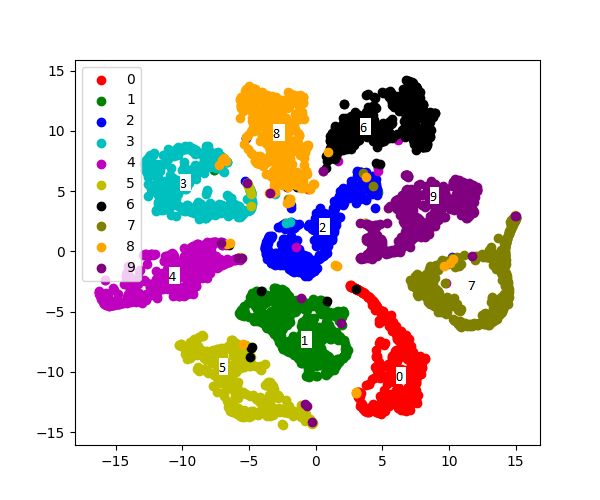}}
		{{\footnotesize (a)}}
	\end{minipage}
	%%\hfill
	\begin{minipage}[b]{0.45\linewidth}
		\centering
		\centerline{\includegraphics[scale=0.4]{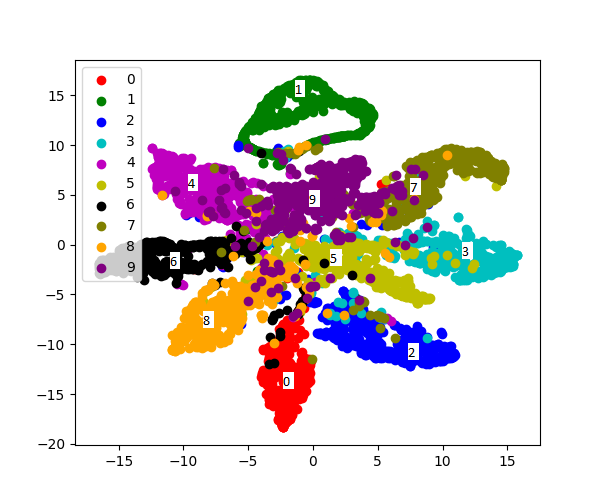}}%, height=5cm
		{{\footnotesize(c)}}
    \end{minipage}
    \begin{minipage}[b]{0.45\linewidth}
		\centering
		\centerline{\includegraphics[scale=0.4]{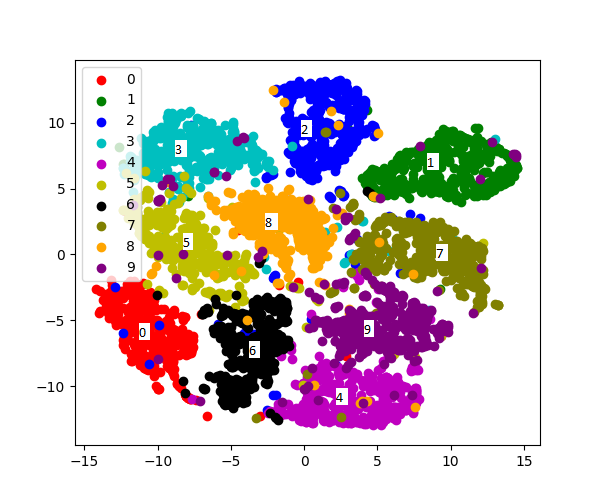}}%, height=5cm
		{{\footnotesize(c)}}
    \end{minipage}
        \begin{minipage}[b]{0.45\linewidth}
		\centering
       \captionsetup{justification=centering}
		\centerline{\includegraphics[scale=0.4]{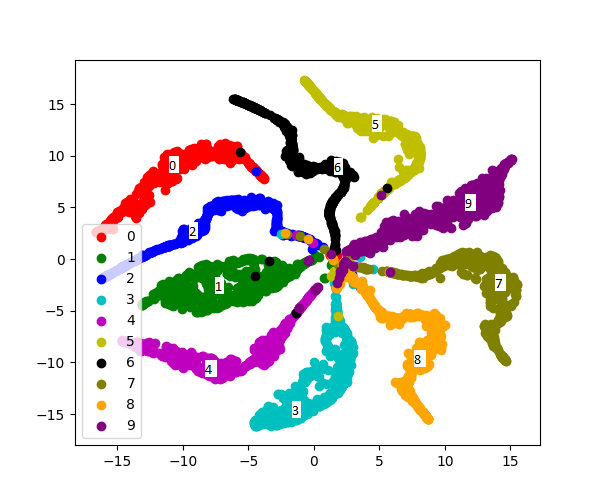}}
		{{\footnotesize (b)}}
	\end{minipage}
	\caption{t-SNE projection \cite{maaten2008visualizing} of the hidden activation of last layer of 4-layer perceptron network using (a) Sigmoid (b) SeLU (c) Tanh (d) ReLU activation function trained on 5000 MNIST handwritten digits test samples. All Networks have 1000 hidden units in all layers}\label{fig444}
\end{figure*}
\indent
The number of redundant features was computed as in \eqref{MyEq12} after the models have been fully trained. Figures~\ref{fig1} a,b,c, and d show the performance of multilayer perceptron with one, two, three, and four hidden layer(s), respectively. The average number of redundant features across all layers of the network is denoted as $\bar{n}_r$. It can be observed in Figure~\ref{fig1} that both width (number of hidden units per layer) and depth (number of layers in the network) increase $\bar{n}_r$. As the number of hidden units per layer increases, $\bar{n}_r$ grows almost linearly. Also, the higher the number of hidden layers in a network, the higher the average number of redundant features extracted and the higher the average feature pairwise correlations. For instance, the network with one hidden layer and $100$ hidden units does not have any feature pair correlated above $0.4$. However, as the depth increases (for two or more hidden layers) more feature pairs have correlation above $0.4$. This observation is similar for other hidden layer sizes ($200$, $300$, $500$, $700$, and $1000$) and depth. In particular, as can be observed in Figure~\ref{fig1}d that many feature pairs in deep multilayer network (with four hidden layers) are almost perfectly correlated with cosine similarity of $0.9$ even with just $100$ hidden units per layer.\\
\indent
In the second set of experiments, we also used the MNIST dataset to see the impact of activation function and number of layers in DNNs on susceptibility to redundant feature extraction. We considered five popular activation functions namely: \emph{Sigmoid}, \emph{Tanh}, \emph{ReLU}, \emph{ELU}, and \emph{SeLU}. In order to focus on the effect of activation function and number of layers, we fixed the width (number of hidden units) of the network for all layers and was set to $1000$. For all networks, the weights were initialized randomly by sampling from normal distribution with zero mean and standard deviation of 0.01. Pairwise feature similarity was measured at thresholds $\tau=0.5$, $0.6$, and $0.7$. As shown in Figure~\ref{fig2} for all thresholds, \emph{sigmoid}, \emph{ELU}, and \emph{SeLU} have higher tendency to extract redundant features than those of \emph{Tanh} and \emph{ReLU}. In fact, it was observed that the average number of redundant feature extracted for \emph{ReLU} and \emph{Tanh} did not increase as the number of layers increased. In fact, as can be observed in Figure~\ref{fig2}b, the redundancy slightly decreases for both \emph{ReLU} and \emph{Tanh} as opposed to \emph{Sigmoid}, \emph{ELU}, and \emph{SeLU} where it is almost always increasing.\\
\indent
Also, Figure~\ref{fig4}c reinforces the observation that ReLU activation is able to outperform both sigmoid and tanh activations in Figures~\ref{fig4}a and b for very deep networks and exhibits inherent nature to extract more diverse features. It can be observed in Figures~\ref{fig4} that \emph{ReLU} benefits from both width and depth than its counterparts. As width and depth increase, the performance of tanh deteriorates while that of sigmoid heavily fluctuates. Deep multilayer network was also evaluated based on the distribution of data in high level feature space. In this regard, t-distributed stochastic neighbor embedding (t-SNE) \cite{maaten2008visualizing} was used to project and visualize the last hidden activations of single-hidden-layer and four-hidden-layer networks using Sigmoid, SeLU, Tanh, and ReLU activations into 2D. The projections of single layer networks and that of four-layer networks are as shown in Figures~\ref{fig44} and \ref{fig444}, respectively. The t-SNE projections show that networks with four hidden layers have clustered activations compared to that of a single layer resulting in within class holes. This is observation is pronounced for Sigmoid and SeLU activations.\\
%We suspect that \emph{tanh} is able to maintain lesser redundancy because of its ability to more extract negatively correlated features due to its ability
%to output both negative and positive values. We remark that the polarity of correlation of two features can be changed by multiplying one of them with a
%negative number. That is, positively-correlated features can easily be translated into negatively-correlated ones in case of \emph{tanh}. However,
%\emph{ReLU} activation does not benefit from this phenomenon since its output is always nonnegative. This may explain why it \\
%\indent
%Another important observation is that
%Each layer has $1024ReLu-activated hidden units and
\indent
In the third set of experiments on MNIST, five weight initialization heuristics were tested and the width of the network per layer was also fixed and set to $1000$. Sigmoid was used in this set of experiments and only the initialization method and number of layers were varied. As shown in Figure~\ref{fig3}, all weight initializations for shallow networks have similar tendency to extract redundant features. As the number of layers increases, however, \emph{he\_normal} \cite{he2015delving} extracts less redundant features than all its counterparts. This observation is relatively consistent for thresholds $\tau=0.5$, $0.6$, and $0.7$ as shown in Figures~\ref{fig3}a, b, and c, respectively. This might explain why it usually outperforms other initialization methods in most vision task.\\
\indent
\begin{figure*}[htb!]
\centering
%\captionsetup{justification=centering,margin=2cm}
	\begin{minipage}[b]{0.32\linewidth}
		\centering
%       \captionsetup{justification=centering}
		\centerline{\includegraphics[scale=0.35]{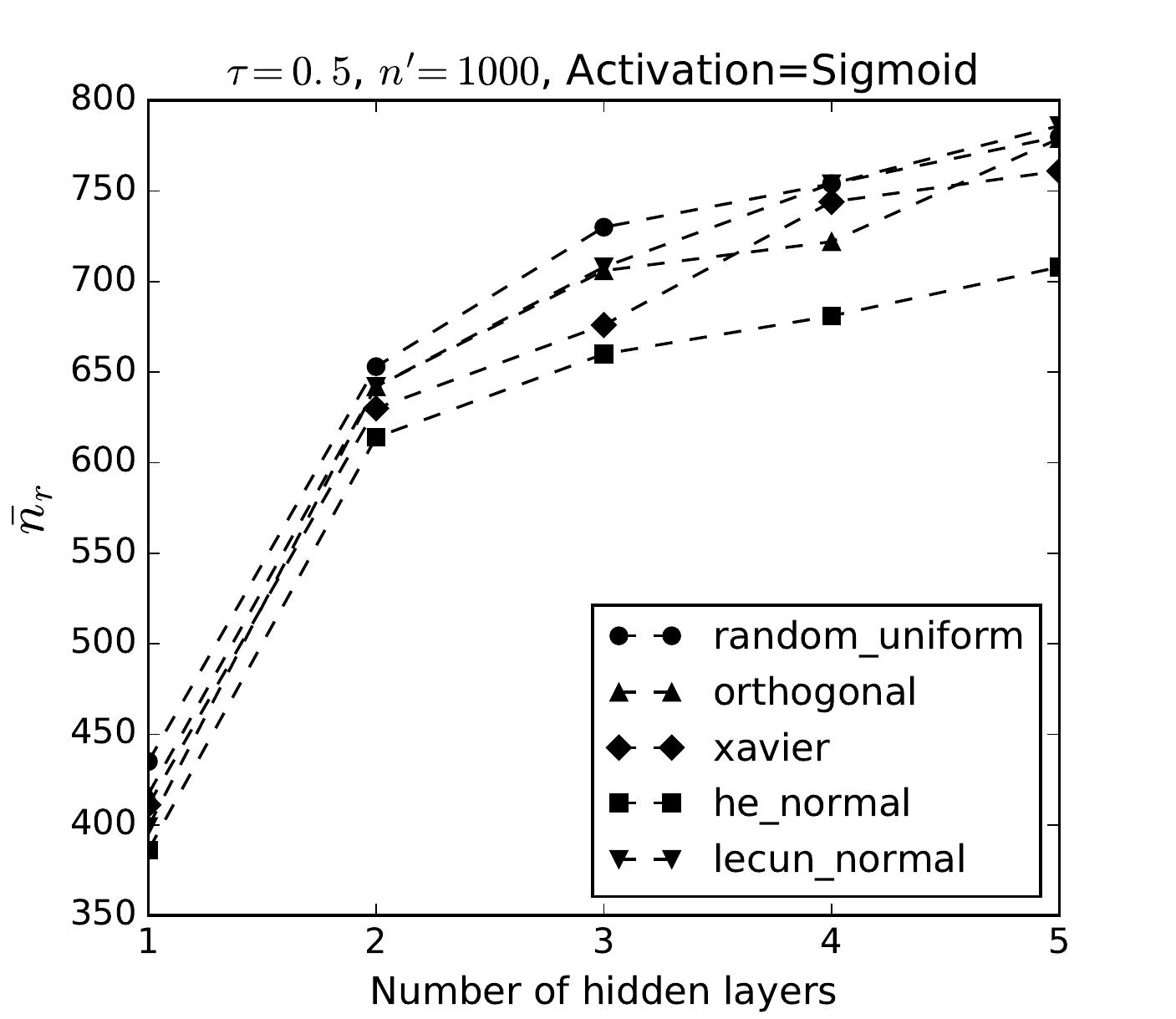}}
		{{\footnotesize (a)}}
	\end{minipage}
	%%\hfill
	\begin{minipage}[b]{0.32\linewidth}
		\centering
		\centerline{\includegraphics[scale=0.35]{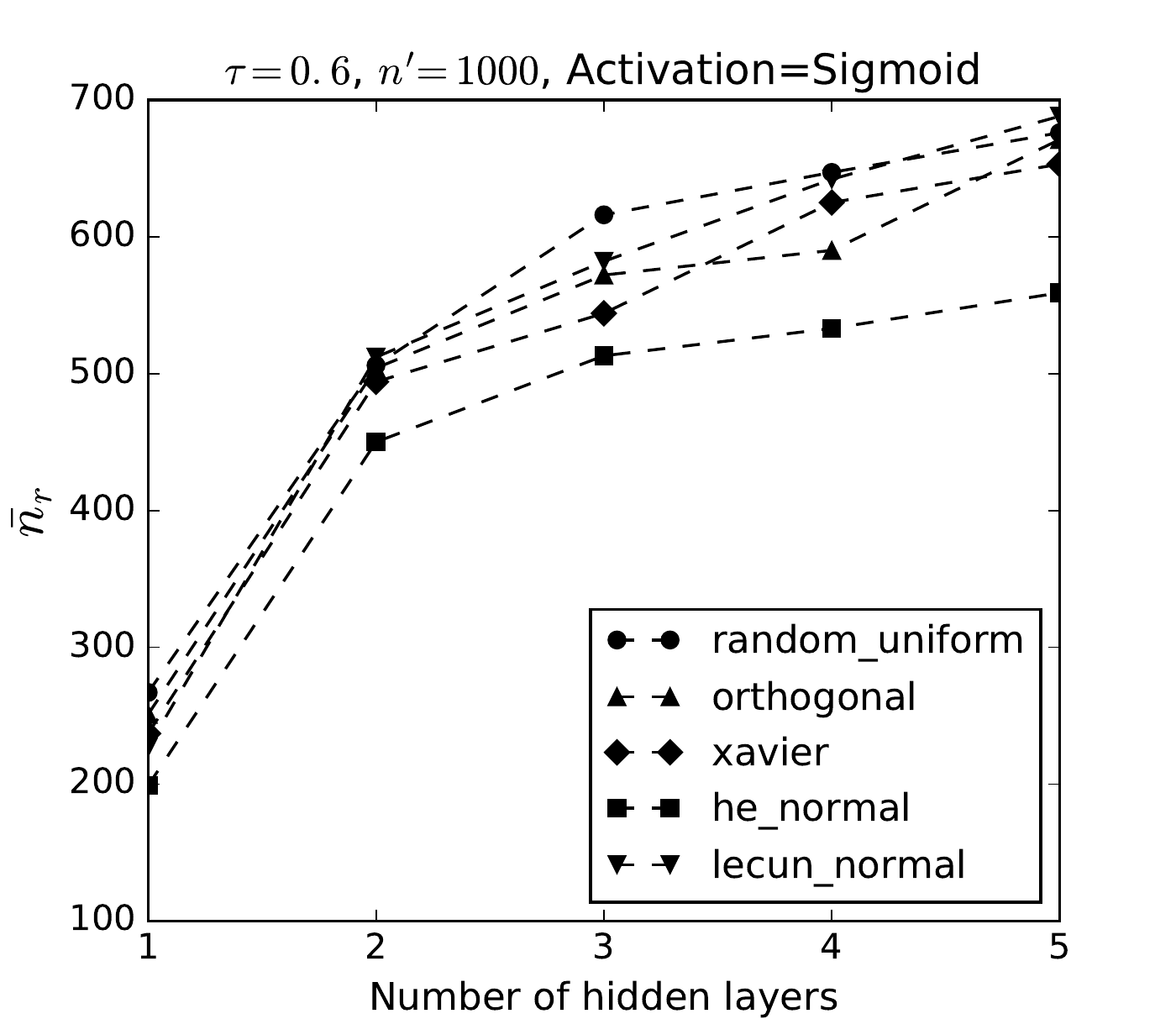}}%, height=5cm
		{{\footnotesize(b)}}
    \end{minipage}
    %%\hfill
	\begin{minipage}[b]{0.32\linewidth}
		\centering
		\centerline{\includegraphics[scale=0.35]{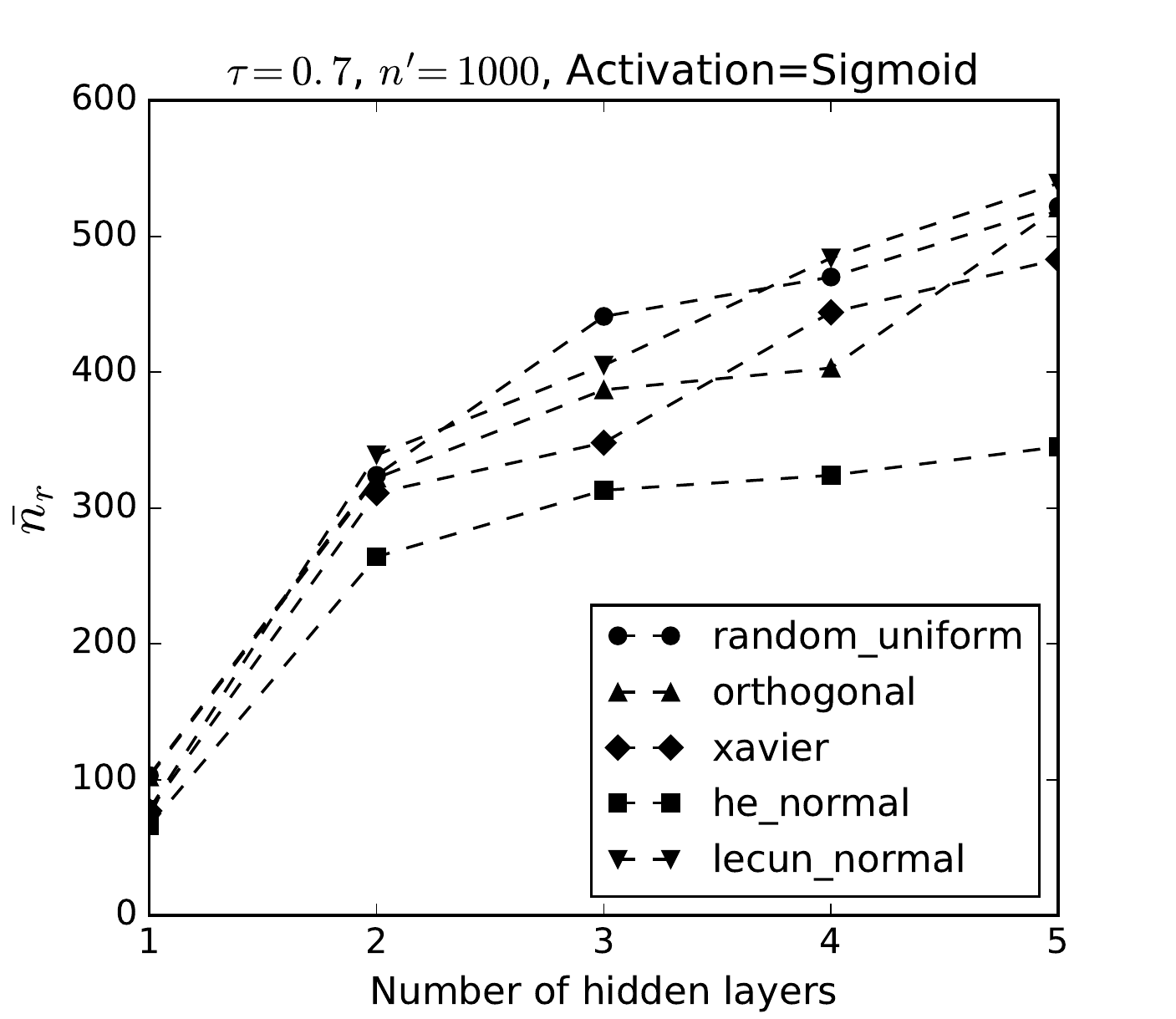}}%, height=5cm
		{{\footnotesize(c)}}
    \end{minipage}
	\caption{Average number of redundant features vs number of hidden layers of multilayer perceptron for five weight initialization methods with (a) $\tau=0.5$ (b) $\tau=0.6$ (c) $\tau=0.7$ using MNIST dataset. Number of hidden units ($n'$) per layer is $1000$ and sigmoid activation is used for all layers.}\label{fig3}		 
\end{figure*}
%\begin{table*}[htb!]
%\centering
%\captionsetup{justification=centering,margin=2cm}
%\caption{Performance of VGG models on Cifar-10 dataset. $n'_{avg}$ is the average number of features for all layers of the model and $\tau$ is the threshold of similarity.}
%\scalebox{1.2}{
%\begin{tabular}{|c|c|c|c|c|c|c|c|} % centered columns (3 columns)
%\hline\hline%inserts double horizontal lines
%\multicolumn{3}{|c|}{VGG} & \multicolumn{4}{c|}{$\bar{n}_r$} & \\ [0.5ex]
%\cline{1-8}
%\multicolumn{1}{|c|} {Model} & \# Conv Layers & $n'_{avg}$ & $\tau=0.4$ & $\tau=0.5$ & $\tau=0.6$ & $\tau=0.7$ & test accuracy (\%)\\
%\hline %\hline %inserts double horizontal lines        	   	   				
%VGG-11 & 8 & 344 & 117 & 83 & 61 & 44 & 92.09\\[.5ex] % inserting body of the table			
%\cline{1-8}
%VGG-13 & 10 & 294 & 111 & 79 & 56 & 38 & 93.65\\[.5ex]
%\cline{1-8}
%VGG-16 & 13 & 325 & 191 & 171 & 148 & 121 & 93.51\\[.5ex]
%\cline{1-8}
%VGG-19 & 16 & 344 & 249 & 235 & 222 & 207 & 93.24\\[.5ex]
%\hline\hline %inserts double horizontal lines
%\end{tabular}
%\label{table:result1}} % is used to refer this table in the text
%\end{table*}
In the last set of experiments, we trained deep convolutional neural networks (VGG-11,13,16,19) on CIFAR-10 dataset to see how depth is impacting redundant feature extraction. VGG architecture \cite{simonyan2014very} is a high capacity network designed originally for ImageNet dataset. We used a modified version of the VGG network architecture, which has $c$ convolutional layers and 2 fully connected layer. Constant $c$ in VGG-11, VGG-13, VGG-16, and VGG-19 are $8$, $10$, $13$, and $16$, respectively. In the modified version of VGG architectures, each layer of convolution is followed by a Batch Normalization layer \cite{ioffe2015batch}. CIFAR-10 dataset contains a labeled set of 60,000 32x32 color images belonging to 10 classes: airplanes, automobiles, birds, cats, deer, dogs, frogs, horses, ships, and trucks. The dataset is split into $50000$ and $10000$ training and testing sets, respectively. Our baseline model was trained for 300 epochs, with a batch-size of 128 and a learning rate 0.1. The learning rate was reduced by a factor of 10 at 150 and 250 epochs. \\
\indent
As shown in Table~\ref{table:result1}, the average number of redundant fetaures across all layers ($\bar{n}_r$) also increases as number of convolutional layers increases. It can be observed that with 13 layers of convolution, the performance of the model starts deteriorating and the percentage of redundant feature increases by more than $21\%$ for all $\tau$ values considered. Another crucial observation is that networks (VGG-11 and VGG-13) with 8 and 10 convolutional layers have relatively similar level of redundancy, especially for $\tau=0.7$. This means, in relative terms, that both VGG-11 and VGG-13 have smaller degree of overfitting compared to VGG-16 and VGG-19 as reflected in their test accuracies. This may suggest a strong correlation between the level of redundancy in a model and its generalization. It must be noted that the test error of VGG-11 is higher than its counterparts; we believe that perhaps VGG-11 with 8 layers of convolution is somehow underfitting the CIFAR-10 dataset.
\begin{table*}[htb!]
\centering
%\captionsetup{justification=centering,margin=2cm}
\caption{Performance of VGG models on Cifar-10 dataset. $\tau$ is the threshold of similarity.}
\scalebox{1.}{
\begin{tabular}{|c|c|c|c|c|c|c|} % centered columns (3 columns)
\hline\hline%inserts double horizontal lines
\multicolumn{2}{|c|}{VGG} & \multicolumn{4}{c|}{$\bar{n}_r$ (\%)} & \\ [0.5ex]
\cline{1-7}
\multicolumn{1}{|c|} {Model} & \# Conv Layers & $\tau=0.4$ & $\tau=0.5$ & $\tau=0.6$ & $\tau=0.7$ & test accuracy (\%)\\
\hline %\hline %inserts double horizontal lines        	   	   				
VGG-11 & 8 & 34.0 & 24.1 & 17.7 & 12.8 & 92.09\\[.5ex] % inserting body of the table			
\cline{1-7}
VGG-13 & 10 & 37.8 & 26.9 & 19.1 & 12.9 & 93.65\\[.5ex]
\cline{1-7}
VGG-16 & 13 & 58.8 & 52.6 & 45.5 & 37.2 & 93.51\\[.5ex]
\cline{1-7}
VGG-19 & 16 & 72.4 & 68.3 & 64.5 & 60.3 & 93.24\\[.5ex]
\hline\hline %inserts double horizontal lines
\end{tabular}
\label{table:result1}} % is used to refer this table in the text
\end{table*}
\section{Conclusion}
This paper shows how size, choice of activation function, and weight initialization impact redundant feature extraction of deep neural network models. The number of redundant features is estimated by agglomerating features in weight space according to a well-defined similarity measure. Experiments were carried out using benchmark datasets and select models. The results show that both width and depth strongly correlate with redundant feature extraction. It is also established that the wider and deeper a network becomes, the higher is its tendency to extract redundant features. It has also been empirically shown on select examples that \emph{ReLU} activation function enforces extraction of less redundant features in comparison with other activations function considered. Also, the \emph{he\_normal} initialization heuristic presented in \cite{he2015delving} offers the advantange of extracting more distinct features for deep networks than other popular initialization heuristics considered. We illustrated the concept using fully-connected and convolutional neural networks on MNIST handwritten digits and CIFAR-10.
\bibliographystyle{IEEEtran}
\bibliography{reference_ijcnn2018}
\end{document}